\documentclass[10.5pt,letterpaper]{article}
\usepackage[top=1.2in,left=1.2in,bottom=1.2in,footskip=0.7in]{geometry}

\usepackage{amsthm}
\usepackage{cite}
\usepackage[pdftex]{graphicx}
\graphicspath{{../pdf/}{../jpeg/}}
\DeclareGraphicsExtensions{.pdf,.jpeg,.png}
\usepackage[cmex10]{amsmath}
\usepackage{algorithmic}
\usepackage{subfigure}
\usepackage{amsfonts}
\usepackage{amsthm}
\usepackage{amssymb}
\usepackage{eufrak}
\usepackage{mathrsfs}
\usepackage{txfonts}
\usepackage{float}

\newtheorem{mythm}{Theorem}

\begin{document}
\begin{centering}
  {\Large
{\textbf{\Large Domain Generalization via Domain-based Covariance Minimization} \\ 
\vspace{0.2in}
Anqi Wu\\
\vspace{0.05in}
{\large Georgia Institute of Technology}\\
\vspace{0.05in}
{\large \hspace{1in} anqiwu.angela@gmail.com}}
}
\newline
\\
\end{centering}



\section*{Abstract}
Researchers have been facing a difficult problem that data generation mechanisms could be influenced by internal or external factors leading to the training and test data with quite different distributions, consequently traditional classification or regression from the training set is unable to achieve satisfying results on test data. In this paper, we address this nontrivial domain generalization problem by finding a central subspace in which domain-based covariance is minimized while the functional relationship is simultaneously maximally preserved. We propose a novel variance measurement for multiple domains so as to minimize the difference between conditional distributions across domains with solid theoretical demonstration and supports, meanwhile, the algorithm preserves the functional relationship via maximizing the variance of conditional expectations given output. Furthermore, we also provide a fast implementation that requires much less computation and smaller memory for large-scale matrix operations, suitable for not only domain generalization but also other kernel-based eigenvalue decompositions. To show the practicality of the proposed method, we compare our methods against some well-known dimension reduction and domain generalization techniques on both synthetic data and real-world applications. We show that for small-scale datasets, we are able to achieve better quantitative results indicating better generalization performance over unseen test datasets. For large-scale problems, the proposed fast implementation maintains the quantitative performance but at a substantially lower computational cost.

\section{Introduction and Related Work}
In the area of machine learning, researchers have been facing a difficult problem that data generation mechanisms could be influenced by internal or external factors leading to the training and test data with quite different distributions, consequently traditional classification or regression from the training set is unable to achieve satisfying results on test data. To illustrate the problem, we consider an example taken from the oil field \cite{liu2011semi}. There are $N$ fields in different locations with multiple wells for each. Experts want to make predictions for each oil production well given historical data that whether it fails to run normally or not. Correctly identifying failure wells is vital for timely repair, reducing oil losses, and saving human, financial, and material resources.
Nevertheless, manually labeling is time-consuming and lacks early warning. To automate well labeling, we aim at adapting models from training fields (source domains) to test fields (target domains). However, given various geographical environments and weather conditions, different fields' data may possess diverse characteristics and distributions, violating the basic assumption that training and test data come from the same distribution.

To solve such problems, a considerable effort has been made in domain adaptation and transfer learning to remedy such problems (\cite{pan2010survey} and its reference therein). The general idea is to transfer useful information from the source domain to improve test accuracy on the target domain. Some directly transfer relevant instances from the source, which assumes that certain parts of the data in the source domain can be reused for learning in the target domain by re-weighting \cite{dai2007boosting, dai2007transferring, quionero2009dataset, jiang2007instance, liao2005logistic}. Some aim at learning a good feature representation for the target domain, in which case, the knowledge used to transfer across domains is encoded into the learned feature representation \cite{raina2007self, dai2007co, ando2005high, blitzer2006domain}. With the new feature representation, the performance of the target task is expected to improve significantly. A third case is to transfer parameters shared by the source domain and the target domain, which the transferred knowledge is encoded into \cite{lawrence2004learning, bonilla2008multi, schwaighofer2004learning, evgeniou2004regularized}. Moreover, there's a paper transferring relational knowledge \cite{mihalkova2007mapping}. 

Nevertheless, the drawback of such approaches is that the target domain should be accessible during the training procedure and is used to enhance models. In addition, this process is repeated for each new target domain, which can be time-consuming. For example, it's commonly seen that a new oil field arrives for failure prediction with thousand of wells, which means training a model for the new field would take a lot of time, and maintaining it results in a large memory occupation. Furthermore, in medical diagnostics, retraining models for every new patient would be unrealistic since time is a valuable asset for diagnosing patients' physical condition.

\begin{figure}[htbp] 
   \centering
   \includegraphics[width=0.7\textwidth]{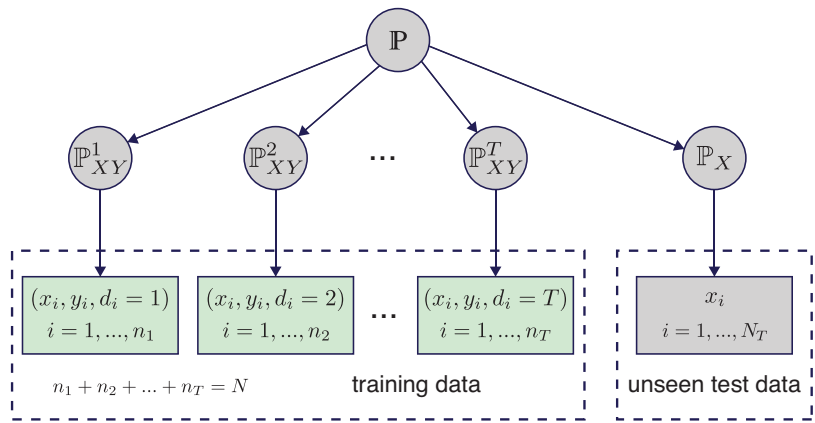} 
   \caption{A simplified schematic diagram of domain generalization adapted from \cite{muandet2013domain}. More notational details are described in Sec. 2 and Sec. 3.}
   \label{framework}
\end{figure}

Accordingly, \cite{muandet2013domain} proposed a framework of domain generalization considering how to take knowledge obtained from an arbitrary number of related source domains and apply it to previously unobserved target domains. Although the marginal distributions $\mathbb{P}(X)$ could vary a lot due to individual factors, the conditional distribution or functional relationship $\mathbb{P}(Y|X)$ is stable across domains, given the fact that oil well-working status always relates to certain production-related attributes and patients' health status is also associated with some measurement indicators. The goal of domain generalization is to estimate a functional relationship that handles changes in the marginal $\mathbb{P}(X)$ or conditional $\mathbb{P}(Y|X)$, shown in Fig.~\ref{framework}.

In paper \cite{muandet2013domain}, the authors introduced Domain Invariant Component Analysis (DICA), a kernel-based algorithm that finds a transformation of the data that (i) minimizes the difference between marginal distributions $\mathbb{P}(X)$ of domains as much as possible while (ii) preserving the functional relationship $\mathbb{P}(Y|X)$. They define a distributional variance for marginal distribution minimization via the inner product of mean maps in kernel space. On the contrary, our work employs a variance measurement based on the covariance form (outer product) of mean maps that matches the general definition of variance with solid theoretical support from Covariance Operator Inverse Regression (COIR) \cite{kim2011central}. The model is named Domain-based Covariance Minimization (DCM). In addition, we have better performances proved by experiment results. Moreover, we show that DCM is closely related to some well-known kernel-based dimension reduction algorithms such as kernel principal component analysis (KPCA) \cite{scholkopf1998nonlinear, muandet2012learning} and COIR, similar to DICA. 

Meanwhile, in many real-world problems, both source and target domains would always have a great amount of data that is usually insurmountable via current domain generalization methods. Our work is eventually equivalent to a large matrix eigenvalue decomposition problem consisting of multiple kernels, which has been studied for a long while. Some state-of-art techniques have tackled similar large-scale kernel matrix problems. For example, for large-scale datasets, KPCA would be very time consuming (with complexity $O(N^3)$, where $N$ is the number of data points) and occupy large memory (to store $N\times N$ kernel matrices), which prohibit it from being used in many applications. To alleviate the computational demand, \cite{rosipal2001expectation} proposed an expectation maximization (EM) approach for performing KPCA by considering KPCA from a probabilistic perspective, and the complexity of their method is $O(kN^2)$ per iteration, where $k$ is the number of extracted components. This comes at the price that a large kernel matrix needs to be stored during the computation. \cite{kim2005iterative} provided a new iterative method, the kernel Hebbian algorithm (KHA), to estimate the kernel principal components with only linear order memory complexity but cannot be used for our problem. \cite{zheng2005improved} presented a ``divide-and-rule" based method that could reduce the computational intensity and memory efficiently, while only suits for eigenvalue decomposition for kernel covariance matrices and is difficult to extend to a more general situation like our expression. For kernel sliced inverse regression (KSIR), \cite{yeh2009nonlinear} mentioned a fast implementation algorithm, especially for the eigenvalue decomposition expression for KSIR, which introduces the Nystrom approximation to approximate the full kernel. However, this is a direct application and is hard to be generalized as well.  

Therefore, we also propose a fast version of DCM, FastDCM, which can deal with such large-scale problems with fast implementations, not only applicable for domain generalization but also some kernel-based dimension reduction algorithms, including the above-mentioned methods. We adopt a low-rank Nystrom approximation to approximate the huge and dense kernel covariance matrix and introduce a fast eigenvalue decomposition technique for estimating effective dimension reduction subspace (e.d.r.) directions in the FastDCM implementation. These reduction techniques will speed up the computation and increase the numerical stability. 

In a nutshell, the novelty of our learning framework is two-fold: 
\vspace{-0.1in}
\begin{itemize}
\item We propose a novel variance measurement for multiple domains to minimize the difference between conditional distributions $\mathbb{P}(X|D)$ ($D$ indicates domains) with solid theoretical demonstration and supports, meanwhile the algorithm preserves the functional relationship $\mathbb{P}(Y|X)$ via maximizing the variance of conditional expectations $\mathbb{E}(X|Y)$. 
\vspace{-0.1in}
\item We can achieve fast computation and small memory for large-scale matrix computation, suitable for not only domain generalization but also other kernel-based eigenvalue decompositions. 
\end{itemize}

\section{Notations}
Let $\mathcal{X}$ denote a nonempty $n-$dimensional input space, $\mathcal{Y}$ an arbitrary output space and $\mathcal{D}$ a discrete domain space. A domain is defined as a joint distribution $\mathbb{P}_{XYD}$ on $\mathcal{X}\times \mathcal{Y}\times \mathcal{D}$ contained in the set $\mathfrak{B}_{\mathcal{X}\times \mathcal{Y}\times \mathcal{D}}$ for all domains. Domains are sampled from a probability distribution $\varmathbb{P}$ on $\mathfrak{B}_{\mathcal{X}\times \mathcal{Y}\times \mathcal{D}}$ with a bounded second moment. 
We assume that $N$ data samples from $T$ domains, $\mathcal{S}=\{x_i,y_i,d_i\}_{i=1}^N$, $d_i\in\{1,2,...,T\}$, are sampled from $\mathbb{P}_{XYD}$. Since $\mathbb{P}(X,Y|D=i)\neq \mathbb{P}(X,Y|D=j)$, the samples in $\mathcal{S}$ are not i.i.d.. (See Fig.~\ref{framework}.)
   
Denote $\mathcal{H}_x$, $\mathcal{H}_y$ and $\mathcal{H}_d$ as reproducing kernel Hilbert spaces (RKHSes) on $\mathcal{X}$, $\mathcal{Y}$ and $\mathcal{D}$ with kernels $k_x:\mathcal{X}\times \mathcal{X}\rightarrow \mathbb{R}$, $k_y:\mathcal{Y}\times \mathcal{Y}\rightarrow \mathbb{R}$ and $k_d:\mathcal{D}\times \mathcal{D}\rightarrow \mathbb{R}$, respectively. The associated mappings are $x\rightarrow \phi(x)\in \mathcal{H}_x$, $y\rightarrow \varphi(y)\in \mathcal{H}_y$ and $d\rightarrow \psi(d)\in \mathcal{H}_d$ induced by the kernels $k_x(\cdot,\cdot)$, $k_y(\cdot,\cdot)$ and $k_d(\cdot,\cdot)$. Without loss of generality, we assume all the feature maps have zeros means, i.e., $\sum_{i=1}^N\phi(x_i)=\sum_{i=1}^N\varphi(y_i)=\sum_{i=1}^N\psi(d_i)=0$. Let $\Sigma_{\alpha\beta}$ be the covariance operator between $\alpha$ and $\beta$ ($\alpha, \beta \in \{X, Y, D\}$) in and between the RKHSes. 

 \section{Domain-based Covariance Minimization}
Domain-based Covariance Minimization (DCM) aims at finding a central subspace in which domain-based covariance is minimized while the functional relationship is maximally preserved. By functional relationship, we only refer to the relationship between input $X$ and actual output $Y$, which is $\mathbb{P}(Y|X)$. 

\subsection{Formation of DCM}
\textbf{Minimizing Domain-based Covariance.}
First, we need to find a domain-based covariance, which aims at extracting directions with the smallest dissimilarity among different domain distributions across all dimensions. Since we always work in RHKS, we drop feature mapping notations for convenience, i.e., $\phi$, $\varphi$, and $\psi$, and assume the dimensionality of the original input space and its RHKS to be $n$. As the previous assumption that the fundamental distribution $\mathbb{P}_{XYD}$ is a joint probability distribution of $X$, $Y$, and $D$, the domain label $D$ is similarly treated as the class label (output $Y$). If different domains distribute separately, then $\mathbb{P}(D|X)$ could vary a lot. Consequently, we want to find an orthogonal transform $B$ onto a low-dimensional ($m-$dimensional) subspace that can least differentiate $\mathbb{P}(D|B^\top X)$. We borrow an intuitive idea from Inverse regression \cite{li1991sliced} to minimize domain-based covariance. We can inversely find $m$ directions minimally telling domain labels apart, which means to find the $m$ smallest eigenvectors of $\mathbb{V}(\mathbb{E}[X|D])$. This is the domain-based covariance we are searching for, and its $m$ smallest eigenvectors denote the directions in which domain distributions exhibit the smallest dissimilarity.

From another perspective, if we want to find an orthogonal transform $B$ preserving the relationship between $X$ and $D$, i.e., $D\bot X|B^\top X$, according to the Inverse Regression framework (see \cite{li1991sliced} for more details and proofs), we assume that $X$ is centered, i.e., $\mathbb{E}(X)=0$ without loss of generality, then
\begin{mythm}
\label{thm1}
If there exists an $m$-dimensional central subspace with bases $B=[\bold{b}_1, ...,\bold{b}_m]\in\mathbb{R}^{n\times m}$, such that $D\bot X|B^\top X$, and for any ${a}\in \mathbb{R}^n$, $\mathbb{E}[{a}^\top X|B^\top X]$ is linear in $\{\bold{b}_l^\top X\}_{l=1}^m$, then $\mathbb{E}(X|D)\in \mathbb{R}^n$ lies on the subspace spanned by $\{\Sigma_{xx}\bold{b}_l\}_{l=1}^m$, where $\Sigma_{xx}\in\mathbb{R}^{n\times n}$ is the covariance of $X=[x_1, x_2, ..., x_N]\in\mathbb{R}^{n\times N}$.
\end{mythm}
In the light of Theorem \ref{thm1}, $B$ can be obtained from the $m$ principal directions of $\mathbb{E}[X|D]$ if aiming at preserving the functional relationship. The other way round, to maximally eliminate the independence between $X$ and $D$ given $B^\top X$, we could find the column vectors of $B$ coinciding with the $m$ smallest eigenvectors of $\mathbb{V}(\mathbb{E}[X|D])$, premultiplied by $\Sigma_{xx}^{-1}$. We generalize the linear transformation $B$ to a nonlinear one $B(X)$ which can also be deemed as $B^\top \Phi_x$ and $\Phi_x=[\phi(x_1), \phi(x_2), ..., \phi(x_N)]$.

Subsequently, in order to obtain $\mathbb{V}(\mathbb{E}[X|D])$, we first assume that domain distributions $\mathbb{P}=\{\mathbb{P}^1_{XY}, \mathbb{P}_{XY}^2, ... ,\mathbb{P}^T_{XY}\}$ are sampled from the distribution $\varmathbb{P}$, and data within each domain $i\in\{1,2,...,T\}$ is generated from $\mathbb{P}^i_{XY}$ over $\mathcal{X}\times \mathcal{Y}$. Each probability distribution is represented as an element in RKHS using the {mean map} \cite{muandet2013domain, berlinet2004reproducing, smola2007hilbert, sriperumbudur2010hilbert}
\begin{equation}
\bold{\mu}_{\mathbb{P}}:=\mu[\mathbb{P}_x]:=\mathbb{E}_{\mathbb{P}_x}[k(x,\cdot)]=\int_{\mathcal{X}}k(x,\cdot)d\mathbb{P}_x.
\end{equation}
Thus, $\mathbb{V}(\mathbb{E}[X|D])$ is the covariance of $X$ given $D$ in RKHS defined as $\mathbb{V}(\mathbb{E}[X|D])=\Psi\Psi^\top$ where $\Psi = [\bold{\mu}_{\mathbb{P}^1}, \bold{\mu}_{\mathbb{P}^2}, ..., \bold{\mu}_{\mathbb{P}^T}]\in\mathbb{R}^{n \times T}$. This is different from \cite{muandet2013domain} which defined a Gram matrix $G$ for distributional variance with $G = \Psi^\top\Psi$. For the reason that in RHKS it's always hard or impossible to derive mean map directly, we use the framework from \cite{kim2011central} to estimate $\mathbb{V}(\mathbb{E}[X|D])$ with conditional covariance operator of $X$ given $D$, which allows a nonlinear central subspace with fewer restrictions on the marginal distribution for $X$ and availability for high-dimensional $D$ by jointly exploiting the kernel operators of both $X$ and $D$.

The \textit{covariance operator} is a natural RKHS extension of the covariance matrix in the original space. For two random vectors ${x}$ and ${d}$ endowed with Hilbert spaces $\mathcal{H}_x$ with $k_x(\cdot,\cdot)$ and $\mathcal{H}_d$ with $k_d(\cdot,\cdot)$ respectively, the \textit{cross covariance} $\Sigma_{dx}\triangleq Cov(\psi(d),\phi(x))$. Furthermore, \textit{conditional covariance operator} of $X$ given $D$, denoted by $\Sigma_{xx|d}$, is defined as:
\begin{equation}
\Sigma_{xx|d}\triangleq \Sigma_{xx}-\Sigma_{xd}\Sigma_{dd}^{-1}\Sigma_{dx}.
\end{equation}
\begin{mythm}
\label{thm2}
For any ${f}\in \mathcal{H}_x$, if there exists ${g}\in \mathcal{H}_d$ such that $\mathbb{E}[{f}(x)|d]={g}(d)$ for almost every ${d}$, then $\Sigma_{xx|d} = \mathbb{E}[\mathbb{V}(X|D)]$.
\end{mythm}
According to Theorem \ref{thm2}, $\Sigma_{xx|d}$ equals the expected conditional variance of $X$ given $D$ under mild conditions (see more proofs and details from \cite{muandet2013domain, fukumizu2004dimensionality}). Therefore, using the well-known \textit{E-V-V-E} identity, 
\begin{equation}
\begin{split}
 \mathbb{V}(\mathbb{E}(X|D))=\mathbb{V}(X)-\mathbb{E}(\mathbb{V}(X|D))=\Sigma_{xx}-\Sigma_{xx|d} =\Sigma_{xd}\Sigma_{dd}^{-1}\Sigma_{dx}.\\
\end{split}
\end{equation}
Given the explicit expression of $\mathbb{V}(\mathbb{E}(X|D))$, the basis $\bold{b}$ is the solution to the eigenvalue problem $\mathbb{V}(\mathbb{E}[X|D])\Sigma_{xx}\bold{b}=\lambda\Sigma_{xx}\bold{b}$. Since we are targeting at finding the $m$ smallest eigenvalues, for each $\bold{b}_l$ the objective function is as follows
\begin{equation}
\label{xd}
\min_{\bold{b}_l\in \mathbb{R}^n}\quad \frac{\bold{b}_l\Sigma_{xx}^{-1}\mathbb{V}(\mathbb{E}(X|D))\Sigma_{xx}\bold{b}_l}{\bold{b}_l^\top \bold{b}_l},
\end{equation}
under the condition that $\bold{b}_l$ is chosen to not be in the span of previously  chosen $\bold{b}_l$.
\vspace{0.1in}

\noindent\textbf{Preserving Functional Relationship.}
To maintain the relationship between input $X$ and real output $Y$, the above framework is directly applied via replacing $D$ with $Y$. Instead of minimizing, we want to find the $m$ largest eigenvectors of $\mathbb{V}(\mathbb{E}[X|Y])$ \cite{muandet2013domain, kim2011central}, written as 
\begin{equation}
\label{xy}
\max_{\bold{b}_l\in \mathbb{R}^n}\quad \frac{\bold{b}_l\Sigma_{xx}^{-1}\mathbb{V}(\mathbb{E}(X|Y))\Sigma_{xx}\bold{b}_l}{\bold{b}_l^\top \bold{b}_l}.
\end{equation}
\vspace{0.1in}

\noindent\textbf{Final Optimization Problem.}
Combining eq.~(\ref{xd}) and eq.~(\ref{xy}) together, we attain the optimization function for DCM which finds the solution $B=[\bold{b}_1, ...,\bold{b}_m]$ for
\begin{equation}
\label{final}
\max_{\bold{b}_l\in \mathbb{R}^n}\quad \frac{\bold{b}_l\Sigma_{xx}^{-1}\Sigma_{xy}\Sigma_{yy}^{-1}\Sigma_{yx}\Sigma_{xx}\bold{b}_l+\bold{b}_l^\top \bold{b}_l}{\bold{b}_l\Sigma_{xx}^{-1}\Sigma_{xd}\Sigma_{dd}^{-1}\Sigma_{dx}\Sigma_{xx}\bold{b}_l+\bold{b}_l^\top \bold{b}_l},
\end{equation}
which is equivalent to 
\begin{equation}
\label{finalmatrix}
\max_{B\in \mathbb{R}^{n\times m}}\quad \frac{tr(B^\top (\Sigma_{xx}^{-1}\Sigma_{xy}\Sigma_{yy}^{-1}\Sigma_{yx}\Sigma_{xx}+I_n)B)}{tr(B^\top (\Sigma_{xx}^{-1}\Sigma_{xd}\Sigma_{dd}^{-1}\Sigma_{dx}\Sigma_{xx}+I_n)B)},
\end{equation}
where $I$ is an identity matrix, $n$ is the dimension of $X$ in RHKS. The numerator enforces $B$ to align with the bases of the central space, maximizing the functional relationship, while the denominator aligns $B$ to minimize the domain difference. $B^\top B$ is used to control the complexity, thereby tightening the generalization bounds.

To solve the optimization, we rewrite eq.~(\ref{finalmatrix}) as
\begin{equation}
\label{l1}
\begin{split}
\max_{B\in \mathbb{R}^{n\times m}} \quad tr(B^\top (\Sigma_{xx}^{-1}\Sigma_{xy}\Sigma_{yy}^{-1}\Sigma_{yx}\Sigma_{xx}+I_n)B)\quad s.t.  \quad tr(B^\top (\Sigma_{xx}^{-1}\Sigma_{xd}\Sigma_{dd}^{-1}\Sigma_{dx}\Sigma_{xx}+I_n)B)=1\\
\end{split}
\end{equation}
which yields Lagrangian 
\begin{equation}
\label{l2}
\begin{split}
\mathcal{L}&=tr(B^\top (\Sigma_{xx}^{-1}\Sigma_{xy}\Sigma_{yy}^{-1}\Sigma_{yx}\Sigma_{xx}+I_n)B)-tr((B^\top (\Sigma_{xx}^{-1}\Sigma_{xd}\Sigma_{dd}^{-1}\Sigma_{dx}\Sigma_{xx}+I_n)B-I_m)\Gamma),
\end{split}
\end{equation}
in respect that eq.~(\ref{finalmatrix}) is invariant to scaling, and $\Gamma$ is a diagonal matrix containing Lagrangian multipliers. 

Given data $\{{x}_i, y_i,d_i\}$, we have the sample estimate of $\widehat{\Sigma}_{xy}=\frac{1}{N}\Phi_x\Phi_y^\top$, where $\Phi_x=[\phi(x_1), \phi(x_2), ..., \phi(x_N)]$ and $\Phi_y=[\varphi(y_1), \varphi(y_2), ..., \varphi(y_N)]$. Therefore, we can rewrite
\begin{equation}
\label{a1}
\begin{split}
\Sigma_{xx}^{-1}\Sigma_{xy}\Sigma_{yy}^{-1}\Sigma_{yx}\Sigma_{xx}&=\left(\frac{1}{N}\Phi_x\Phi_x^\top\right)^{-1}\left(\frac{1}{N}\Phi_x\Phi_y^\top\right)\left(\frac{1}{N}\Phi_y\Phi_y^\top +\epsilon I_n\right)^{-1}\left(\frac{1}{N}\Phi_y\Phi_x^\top\right)\left(\frac{1}{N}\Phi_x\Phi_x^\top\right)\\
&=\frac{1}{N}\Phi_x\left(\Phi_x^\top\Phi_x\right)^{-1}\left(\Phi_y^\top\Phi_y\right)\left(\Phi_y^\top\Phi_y+N\epsilon I_n\right)^{-1}\left(\Phi_x^\top\Phi_x\right)\Phi_x^\top.\\
\end{split}
\end{equation}
Let $K_x=\Phi_x^\top\Phi_x\in\mathbb{R}^{N\times N}$, $K_y=\Phi_y^\top\Phi_y\in\mathbb{R}^{N\times N}$, and also let $\bold{b}_l=\sum_{i=1}^N\beta_l^i\phi(x_i)=\Phi_x\bold{\beta}_l$ be the $l^{th}$ basis function of $B$ and $\bold{\beta}_l$ is an $N$-dimensional coefficient vector. Let $\mathcal{B}=[\bold{\beta}_1, \bold{\beta}_2, ..., \bold{\beta}_m]$, then $B=\Phi_x\mathcal{B}$ that is the actual orthogonal transformation we use. In eq.~(\ref{a1}), we use the fact that $\Phi_y^\top\left(\Phi_y\Phi_y^\top +\epsilon I_n\right)^{-1}=\left(\Phi_y^\top\Phi_y +\epsilon I_N\right)^{-1}\Phi_y^\top$. We apply the same rule to the domain related terms, ultimately leading to the Lagrangian form as
\begin{equation}
\label{l2}
\begin{split}
\mathcal{L}=tr(\mathcal{B}^\top (K_y(K_y+N\epsilon I_N)^{-1}K_xK_x+K_x)\mathcal{B})-tr((\mathcal{B}^\top (K_d(K_d+N\epsilon I_N)^{-1}K_xK_x+K_x)\mathcal{B}-I)\Gamma).
\end{split}
\end{equation}
Taking the derivative of eq.~(\ref{l2}) w.r.t $\mathcal{B}$ and setting it to zero, we finally obtain a generalized eigenvalue decomposition problem
\begin{equation}
\label{eigen}
(K_y(K_y+N\epsilon I_N)^{-1}K_xK_x+K_x)\mathcal{B}=(K_d(K_d+N\epsilon I_N)^{-1}K_xK_x+K_x)\mathcal{B}\Gamma.
\end{equation}
The $m$ largest eigenvectors of eq.~(\ref{eigen}) correspond to the transformation $\mathcal{B}$. Table \ref{table:DCM} briefly describes the framework of DCM.
\begin{table}
\centering
\caption{ Algorithm description for DCM}
\label{table:DCM}
\begin{tabular}{p{12cm}}
\hline
\textbf{Algorithm 1} DCM\\
\hline
\textbf{Input}:\quad Parameters $\epsilon, m$, TrainData $X=\{{x}_i, y_i, d_i\}_{i=1}^N$, TestData $Z=\{{z}_i\}_{i=1}^{N_T}$\\
\textbf{Output}:\quad Projection $\mathcal{B}\in \mathbb{R}^{N\times m}$, Projected TrainData $\tilde{K}_x\in \mathbb{R}^{m\times N}$, Projected TestData $\tilde{K}_z\in \mathbb{R}^{m\times N_T}$\\
1. Calculate kernel matrices $K_x(i,j)=k(x_i,x_j)$, $K_y(i,j)=k(y_i,y_j)$, $K_d(i,j)=k(d_i,d_j)$, $K_z(i,j)=k(x_i,z_j)$.\\
2. Solve the generalized eigenvalue problem: $(K_y(K_y+N\epsilon I_N)^{-1}K_xK_x+K_x)\mathcal{B}=(K_d(K_d+N\epsilon I_N)^{-1}K_xK_x+K_x)\mathcal{B}\Gamma$ for $\mathcal{B}$.\\
3. Output $\mathcal{B}$, $\tilde{K}_x=\mathcal{B}^\top K_x$ and $\tilde{K}_z=\mathcal{B}^\top K_z$.\\
\hline
\end{tabular}
\end{table}
\\
\vspace{0.1in}

\noindent\textbf{Relations to Other Approaches.} DCM could be generalized to many supervised and unsupervised dimension reduction techniques. 
\vspace{-0.1in}
\begin{itemize}
\item When there's only one distribution across domains and the minimization of domain covariance could be ignored, i.e., $K_d = 0$, DCM degenerates into covariance operator inverse regression (COIR).
\vspace{-0.1in}
\item In the unsupervised situation with one distribution that is $K_d = 0$ and $K_y = \bold{1}$ (the matrix with all ones), the algorithm would further degenerate into kernel principle component analysis (KPCA).
\end{itemize}

\section{Fast Domain-based Covariance Minimization (FastDCM)}
When source domains have large-scale datasets, derivation of eq.~(\ref{eigen}) would be prohibitive due to the scale of multiple kernels and standard algorithms for computing the eigenvalue decomposition of a dense $N\times N$ matrix take $O(N^3)$ time. We propose a low-rank matrix approximation based eigen-decomposition method to effectively solve the large-scale problem that selects a subset of $M\ll N$ columns from the kernel matrix and subsequently decreases the computation to the scale of $M$ with $O(M^2N)$ time complexity meanwhile avoiding the storage of large kernel matrices. Our FastDCM framework can not only effectively provide a fast implementation for DCM, but is also applicable to large-scale KPCA, DICA, COIR, and all the other kernel-based eigenvalue decomposition problems.

\subsection{Formation of FastDCM}
A general low-rank matrix approximation is the Nystrom method \cite{drineas2005nystrom,fowlkes2004spectral,williams2001using}. It approximates a symmetric positive semidefinite (p.s.d.) matrix $G\in \mathbb{R}^{N\times N}$ by a sample $C$ of $M\ll N$ columns from $G$. Typically, this subset of columns is randomly selected by uniform sampling without replacement \cite{williams2001using, kumar2009sampling}. After selecting $C$, the rows and columns of $G$ can be rearranged as follows

\begin{equation}
C=\left[ \begin{array}{c}
W\\
A\end{array}\right]\quad \mbox{and}\quad 
G=\left[ \begin{array}{c}
W \\
A \end{array}
\begin{array}{c}
A^\top\\
Q\end{array}\right],
\end{equation}
where $W\in \mathbb{R}^{M\times M}$ is symmetric, $A\in \mathbb{R}^{(N-M)\times M}$ and $Q\in \mathbb{R}^{(N-M)\times (N-M)}$. Let's assume the SVD of $W$ is $U\Lambda U^\top$, where $U$ is an orthonormal matrix and $\Lambda=diag(\sigma_1, ..., \sigma_M)$ is a diagonal matrix containing singular values arranged in a non-increasing order. For any $k\leq M$, the rank-$k$ Nystrom approximation is 
\begin{equation}
\tilde{G}_k=CW_k^+C^\top,
\end{equation}
where $W_k^+=\sum_{i=1}^k\sigma_i^{-1}U^{(i)}U^{(i)\top}$, and $U^{(i)}$ is the $i$th column of $U$. If we set $k=M$, then we can approximate $G$ as
 \begin{equation}
\tilde{G}=CW^{-1}C^\top.
\end{equation}
The time complexity is $O(M^2N)$ which is much lower than $O(N^3)$.

Based on such an approximation method, we can replace $K_\alpha$ with $C_\alpha W_\alpha^{-1}C_\alpha^\top$ and let $S_{\alpha \beta}=C_{\alpha}^\top C_{\beta}\in \mathbb{R}^{M\times M}$ ($\alpha, \beta\in \{x,y,d\}$), $\tilde{W}=W^{-1}$. Then the derivation of a standard eigenvalue decomposition version of eq.~(\ref{eigen}) with a low-rank Nystrom approximation presents as follows
\begin{enumerate}
\item We first approximate the left and the right of eq.~(\ref{eigen}) with low-rank matrices
\begin{equation}
\label{derivationofnystrom}
\begin{split}
Left:\\
&(K_y(K_y+N\epsilon I_N)^{-1}K_xK_x+K_x)\mathcal{B}\\
&=(C_y\tilde{W_y}C_y^\top(C_y\tilde{W_y}C_y^\top+N\epsilon I_N)^{-1}C_x\tilde{W_x}C_x^\top C_x\tilde{W_x}C_x^\top+C_x\tilde{W_x}C_x^\top)\mathcal{B}\\
&=(C_y\tilde{W_y}(C_y^\top C_y\tilde{W_y}+N\epsilon I_M)^{-1}C_y^\top C_x\tilde{W_x}C_x^\top C_x\tilde{W_x}+C_x\tilde{W_x})C_x^\top\mathcal{B}\\
&=(C_y\tilde{W_y}(S_{yy}\tilde{W_y}+N\epsilon I_M)^{-1}S_{yx}\tilde{W_x}S_{xx}\tilde{W_x}+C_x\tilde{W_x})C_x^\top\mathcal{B};\\
Right:\\
&(K_d(K_d+N\epsilon I_N)^{-1}K_xK_x+K_x)\mathcal{B}\\
&=(C_d\tilde{W_d}(S_{dd}\tilde{W_d}+N\epsilon I_M)^{-1}S_{dx}\tilde{W_x}S_{xx}\tilde{W_x}+C_x\tilde{W_x})C_x^\top\mathcal{B}.\\
\end{split}
\end{equation}
\item To make sure that we can finally solve the problem with a form of $C_x\Omega C_x^\top$, where $\Omega \in \mathbb{R}^{M\times M}$ is an arbitrary square matrix, we multiply an invertible $K_x$ to both sides of eq.~(\ref{eigen}) which still guarantees the eigenvalues unchanged. Then assuming the invertibility of the right-hand-side matrix in the generalized eigenvalue problem, we transform eq.~(\ref{eigen}) to a standard eigenvalue problem. 
\begin{equation}
\label{derivationofnystrom}
\begin{split}
&(C_x\tilde{W_x}C_x^\top(C_d\tilde{W_d}(S_{dd}\tilde{W_d}+N\epsilon I_M)^{-1}S_{dx}\tilde{W_x}S_{xx}\tilde{W_x}+C_x\tilde{W_x})C_x^\top+N\epsilon I_M)^{-1}\\
&C_x\tilde{W_x}C_x^\top (C_y\tilde{W_y}(S_{yy}\tilde{W_y}+N\epsilon I_M)^{-1}S_{yx}\tilde{W_x}S_{xx}\tilde{W_x}+C_x\tilde{W_x})C_x^\top\\
&=C_x(\tilde{W_x}S_{xd}\tilde{W_d}(S_{dd}\tilde{W_d}+N\epsilon I_M)^{-1}S_{dx}\tilde{W_x}S_{xx}\tilde{W_x}S_{xx}+\tilde{W_x}S_{xx}\tilde{W_x}S_{xx}+N\epsilon I_M)^{-1}\\
&(\tilde{W_x}S_{xy}\tilde{W_y}(S_{yy}\tilde{W_y}+N\epsilon I_M)^{-1}S_{yx}\tilde{W_x}S_{xx}\tilde{W_x}+\tilde{W_x}S_{xx}\tilde{W_x})C_x^\top.\\
\end{split}
\end{equation}
Let 
\begin{equation}
\label{omega}
\begin{split}
\Omega=&(\tilde{W_x}S_{xd}\tilde{W_d}(S_{dd}\tilde{W_d}+N\epsilon I_M)^{-1}S_{dx}\tilde{W_x}S_{xx}\tilde{W_x}S_{xx}+\tilde{W_x}S_{xx}\tilde{W_x}S_{xx}+N\epsilon I_M)^{-1}\\
&(\tilde{W_x}S_{xy}\tilde{W_y}(S_{yy}\tilde{W_y}+N\epsilon I_M)^{-1}S_{yx}\tilde{W_x}S_{xx}\tilde{W_x}+\tilde{W_x}S_{xx}\tilde{W_x})\in \mathbb{R}^{M\times M}.
\end{split}
\end{equation}
Accordingly,  eq.~$(\ref{eigen}) \Leftrightarrow C_x\Omega C_x^\top\mathcal{B}=\mathcal{B}\Gamma$.
\item Up to this point, we already avoid the large kernel matrix storage, inversion, and multiplication. Next, let $Q=C_x\Omega C_x^\top\in \mathbb{R}^{N\times N}$, the ultimate task is to perform eigen-decomposition for matrix $Q$, which is still a large matrix requiring $O(N^3)$ time. While there are many fast eigen-decomposition algorithms, we would demonstrate a more intuitive and natural way for our problems. Based upon the observation of $Q$ with $rank\leq M$, we claim that the eigenvalues of $Q$ would have at most $M$ non-zeros. Thus only $M$ columns of $\mathcal{B}$ should be valid, and there's no need to derive the full matrix of $\mathcal{B}$. 

According to SVD definition from \cite{press1992numerical}, we assume the SVD of $C_x$ is $U\Lambda V^\top$, where $U\in \mathbb{R}^{N\times M}$, $\Lambda\in \mathbb{R}^{M\times M}$, $V\in \mathbb{R}^{M\times M}$, $U^\top U=I_M$, $V^\top V=I_M$ and $\Lambda=diag(\sigma_1, ..., \sigma_M)$ is a diagonal matrix containing singular values arranged in a non-increasing order.
\begin{equation}
\label{qq}
\begin{split}
Q=C_x\Omega C_x^\top=U\Lambda V^\top \Omega V\Lambda U^\top
\Leftrightarrow
QU\Lambda=U\Lambda V^\top \Omega V\Lambda^2.\\
\end{split}
\end{equation}
Since $S_{xx}=C_x^\top C_x=V\Lambda^2V^\top\in \mathbb{R}^{M\times M}$, $V$ is the eigenvector matrix of $S_{xx}$ and can be fast calculated. Subsequently, take $\Lambda^2=V^\top S_{xx}V$ into eq.~(\ref{qq}), we obtain
\begin{equation}
\label{qq1}
\begin{split}
QU\Lambda=U\Lambda V^\top \Omega VV^\top S_{xx}V.\\
\end{split}
\end{equation}
Suppose that the eigenvector and eigenvalue matrices of $V^\top \Omega VV^\top S_{xx}V$ are respectively $\Theta\in \mathbb{R}^{M\times M}$ and $\Delta\in \mathbb{R}^{M\times M}$, then
\begin{equation}
\label{qq2}
\begin{split}
QU\Lambda\Theta=U\Lambda V^\top \Omega VV^\top S_{xx}V\Theta=U\Lambda \Theta\Delta.\\
\end{split}
\end{equation}
Thereby, $\tilde{\mathcal{B}}=U\Lambda\Theta=C_xV\Theta\in \mathbb{R}^{N\times M}$ which corresponds to the $M$ eigenvectors from $\mathcal{B}$ with non-zero eigenvalues and $\tilde{\Gamma}=\Delta\in \mathbb{R}^{M\times M}$ containing $M$ non-zero eigenvalues on the diagonal of $\Gamma$. To ensure the presence of the following component analysis, we should guarantee that $m\leq M$.
\end{enumerate}
\begin{table}
\centering
\caption{Algorithm description for FastDCM}
\label{fastDCM}
\begin{tabular}{p{12cm}}
\hline
\textbf{Algorithm 2} FastDCM\\
\hline
\textbf{Input}:\quad Parameters $\epsilon, m$, TrainData $X=\{{x}_i, y_i, d_i\}_{i=1}^N$, TestData $Z=\{{z}_i\}_{i=1}^{N_T}$\\
\textbf{Output}:\quad Projection $\mathcal{B}\in \mathbb{R}^{N\times m}$, Projected TrainData $\tilde{K}_x\in \mathbb{R}^{m\times N}$, Projected TestData $\tilde{K}_z\in \mathbb{R}^{m\times N_T}$\\
1. Uniformly randomly pick $M$ samples from $X$ to form a selected train set $\tilde{X}$.\\
2. Calculate $C_x(i,j)=k(x_i,\tilde{x}_j)$, $C_y(i,j)=k(y_i,\tilde{y}_j)$, $C_d(i,j)=k(d_i,\tilde{d}_j)$, $K_z(i,j)=k(\tilde{x}_i,z_j)$, and also $\tilde{W}_x$, $\tilde{W}_y$, $\tilde{W}_d$.\\
3. Calculate $\Omega$ according to eq.~(\ref{omega}).\\
4. Eigen decompose $S_{xx}$ to obtain its eigenvector matrix $V$.\\
5. Eigen decompose $V^\top \Omega VV^\top S_{xx}V$ to obtain its eigenvector $\Theta$.\\
6. Output $\tilde{\mathcal{B}} = C_xV\Theta$, $\tilde{K}_x=\tilde{\mathcal{B}}^\top C_x\tilde{W_x}C_x^\top$ and $\tilde{K}_z=\tilde{\mathcal{B}}^\top K_z$.\\
\hline
\end{tabular}
\end{table}
In a nutshell, the framework of FastDCM (Table \ref{fastDCM}) can be widely applied to any kernel-based eigenvalue problems through the following steps:
\vspace{-0.1in}
\begin{itemize}
\item Approximate each kernel matrix with the Nystrom approximation.
\vspace{-0.1in}
\item Reform the original problem to the form of $C_x\Omega C_x^\top \mathcal{B}=\mathcal{B}\Gamma$.
\vspace{-0.1in}
\item Solve the large-scale eigenvalue problem via step (3).
\end{itemize}

\section{Experiments}
We evaluate DCM against some well-known dimension reduction and domain generalization techniques on synthetic data and real-world applications, covering one regression task and four anomaly detection tasks with imbalanced distributions. For the small-scale computation, the major algorithms we compare against are Covariance Operator Inverse Regression (COIR) \cite{kim2011central}, Domain Invariant Component Analysis (DICA) \cite{muandet2013domain} and simple SVM, which takes the original training data and test data as inputs without dimension reduction. In the large-scale situation, we explore a fast version of COIR and DCM that are FastCOIR and FastDCM respectively. Standard versions may be applied on large data and achieve a little bit better results than fast implementations, whereas they sacrifice enormous time and memories, which is inapposite and infeasible for most large datasets. Consequently, we only illustrate the differences with simple SVM.

\subsection{Synthetic Data}
To simulate different marginal distributions, we generate 10 groups of $n_i\sim Poisson(100)$ data points, and each collection is sampled from a 10-dimensional zero-mean Gaussian distribution with covariance generated from Wishart distribution $\mathcal{W}(\eta\times I_{10},10)$, where $\eta$ is an alterable parameter. We set $\eta$ to be 0.1, 0.2, and 0.5 in the experiment. We also keep the functional relationship $y = sgn(sgn(\bold{b}_1^\top \bold{x}+\varepsilon_1)log(|\bold{b}_2^\top \bold{x}+\varepsilon_2+c)))$ consistent for all collections, where $\bold{b}_1$ and $\bold{b}_2$ are weight vectors, $c$ is a constant and $\varepsilon_1, \varepsilon_2\sim \mathcal{N}(0,1)$. 

We use SVM for the classification purpose with a Gaussian RBF kernel for $X$ with parameters $\epsilon=10^{-3}$ and changeable $\gamma=\frac{1}{2\sigma_x^2}$ ($\gamma=0.1, 0.5$ and $\sigma_x$ is the lengthscale of the kernel). For discrete $Y$ and $D$, the kernel matrix is $k(a,b)=\delta(a,b)$ where $\delta(a,b)=1$ if $a=b$. For the following datasets, we adopt the same experimental setting for discrete outputs. We split 10 collections into 7 for training and 3 for test. Due to the small size of synthetic data, we only compare standard algorithms here without fast implementations. Table \ref{synthetictable} reports the average accuracies and their standard deviation over 20 repetitions of the experiments by varying $\eta$ and $\gamma$.

\begin{table}[!t]
\centering
\caption{Classification accuracy on Synthetic Dataset}
\vspace{0.1in}
\label{synthetictable}
\begin{tabular}{|c|c|c|c|c|}
\hline
Datasets &SVM & COIR & DICA & DCM \\
\hline
$\eta=0.1$, $\gamma=0.1$&$69.96\pm5.81$ &$69.89\pm5.27$ &$69.89\pm5.30$ &$\bold{70.31\pm4.91}$ \\
$\eta=0.5$, $\gamma=0.1$&$71.51\pm4.12$ &$71.80\pm4.75$ &$71.82\pm4.80$ &$\bold{72.46\pm4.21}$ \\
$\eta=0.1$, $\gamma=0.5$&$64.87\pm6.81$ &$66.12\pm5.94$ &$66.10\pm5.94$ &$\bold{66.76\pm6.11}$ \\
$\eta=0.2$, $\gamma=0.5$&$59.63\pm4.00$ &$62.58\pm4.06$ &$62.58\pm4.07$ &$\bold{63.03\pm4.46}$ \\
$\eta=0.5$, $\gamma=0.5$&$53.44\pm3.63$ &$57.61\pm3.87$ &$57.61\pm3.87$ &$\bold{59.06\pm4.25}$ \\
\hline
\end{tabular}
\end{table}

\begin{figure}[!t]
  \centering
  \subfigure[COIR]{
    \label{fig:subfig:a} 
    \includegraphics[width=5in]{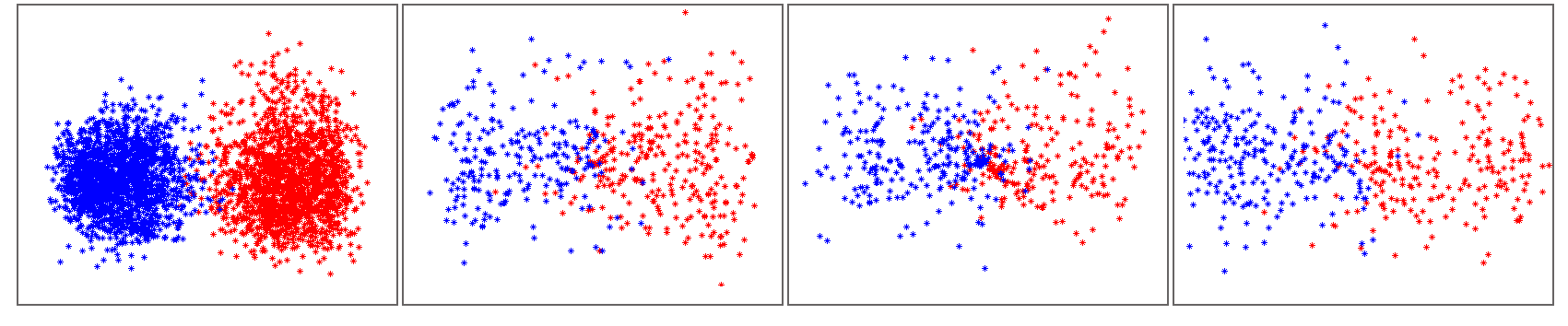}}
  \subfigure[DICA]{
    \label{fig:subfig:b} 
    \includegraphics[width=5in]{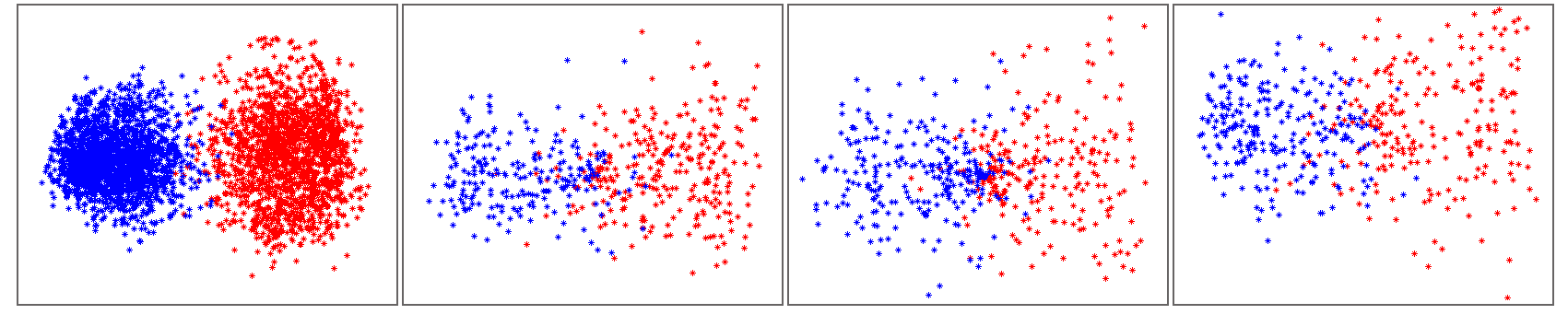}}
  \label{fig:subfig} 
\subfigure[DCM]{
    \label{fig:subfig:c} 
    \includegraphics[width=5in]{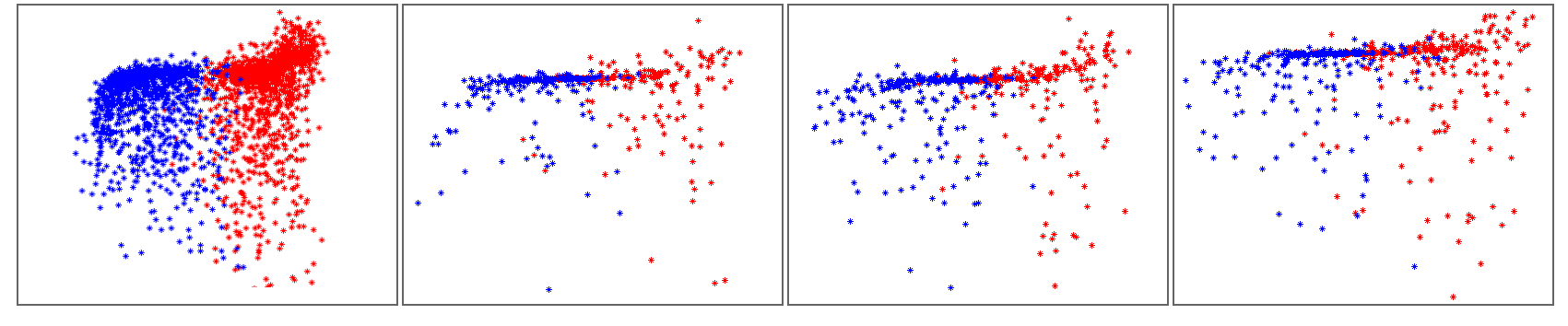}}
    \caption{Projections of the synthetic data onto the first two eigenvectors from COIR, DICA, and DCM, respectively. The leftmost column contains projected training set and the three rightmost columns correspond to three domains of test sets after projection. }
  \label{syntheticfig} 
\end{figure}

Fig.~\ref{syntheticfig} shows the projections of the synthetic data onto the first two eigenvectors from COIR, DICA, and DCM. To demonstrate the projection more clearly, $n_i$ is generated from $Poisson(500)$ and the dimension of Gaussian distribution is reduced to 5 so that distributions don't variate too much and concentrate more to have explicit shapes, and $\gamma=1$. 

\subsection{Parkinson's Telemonitoring Data}
Parkinson's Telemonitoring dataset\footnote{http://archive.ics.uci.edu/ml/datasets/Parkinsons+Telemonitoring} \cite{tsanas2009accurate} is composed of a range of biomedical voice measurements from 42 people with early-stage Parkinson's disease recruited to a six-month trial of a telemonitoring device for remote symptom progression monitoring. The recordings were automatically captured in the patient's homes. The objective is to predict the clinician's Parkinson's disease symptom motor and total scores on the UPDRS scale from 16 voice measures. There are around 200 recordings per patient, and there are roughly 9,000 data points in total. 

\begin{table}[!t]
\centering
\caption{Root Mean Square Error (RMSE) for SVR on Parkinson's Telemonitoring Dataset}
\vspace{0.1in}
\label{parkinsontable}
\begin{tabular}{|c|c|c|c|}
\hline
Methods&Motor Score&Total Score&Time ($s$)\\
\hline
SVR&$9.03\pm0.84$ & $11.41\pm1.39$ & \\
COIR&$9.42\pm0.79$ & $12.23\pm1.46$ & $218.75$  \\
DICA&$9.07\pm0.79$ & $12.35\pm1.48$ & $408.25$ \\
DCM&$\bold{8.61\pm0.98}$ & $\bold{10.71\pm1.53}$ & $101.96$ \\
FastCOIR&$9.13\pm0.79$ & $11.68\pm1.46$ & $9.41$ \\
FastDCM&$\bold{8.65\pm0.91}$ & $\bold{10.76\pm1.52}$ & $11.52$ \\
\hline
\end{tabular}
\end{table}
\begin{table}[!t]
\centering
\caption{AUC and G-Mean for SVM classification on GvHD Dataset}
\vspace{0.1in}
\label{gvhdtable}
\begin{tabular}{|c|c|c|c|}
\hline
Methods ($n_i=200$)&AUC &G-Mean&Time ($s$)\\
\hline
SVM&$0.8904\pm 0.0161$ & $0.8869\pm 0.0175 $&\\
COIR&$0.9058\pm 0.0134$ &  $0.9038\pm 0.0143$&49.84\\
DICA&$0.9059\pm 0.0136$  &$0.9039\pm 0.0145$&133.34\\
DCM&$\bold{0.9065\pm 	0.0136}$ & $\bold{0.9046\pm 0.0146}$&57.76\\
\hline
Methods ($n_i=1000$)&AUC &G-Mean&Time ($s$)\\
\hline
SVM&$0.9020\pm 0.0178$ &$0.8992\pm 0.0196$ & \\
FastCOIR&$0.9035\pm 	0.0147$ &$0.9013\pm 0.0160$ &91.44 \\
FastDCM&$\bold{0.9043\pm 	0.0145}$ &$\bold{0.9021\pm 0.0158}$ &130.59\\
\hline
\end{tabular}
\end{table}

Given that Parkinson's Telemonitoring dataset is a regression problem with continuous output, we directly calculate the kernel matrix for $Y$ with an RBF kernel. The variance $\gamma=\frac{1}{2\sigma_x^2}$ and $\gamma_y=\frac{1}{2\sigma_y^2}$ are both set to be the median of motor and total UPDRS scores. Among 42 patients, 29 are for training, and 13 are for test. Besides, $\varepsilon$-SVR is adopted for the regression purpose, denoted as SVR. Moreover, we apply both standard and fast versions to the dataset for a comprehensive comparison. Table \ref{parkinsontable} shows the root mean square error (RMSE) of motor and total UPDRS scores and running time over 20 repetitions with parameters $\epsilon=10^{-4}$, $M=5$.

\subsection{Gating of Flow Cytometry Data}
Acute graft-versus-host disease (aGVHD) occurs in allogeneic hematopoietic stem cell transplant (SCT) recipients when donor-immune cells in the graft initiate an attack on the skin, gut, liver, and other tissues of the recipient. It has great significance in the field of allogeneic blood and marrow transplantation. \cite{brinkman2007high} found that an increase in the proportion of $CD3^+CD4^+CD8\beta^+$ cells 7-21 days post-transplant best correlated with the subsequent development of aGVHD. Thus our goal is to identify $CD3^+CD4^+CD8\beta^+$. The GvHD dataset consists of 31 patients following allogeneic blood and marrow transplant, while only 30 are used in the experiment due to the insufficiency of one patient. The raw dataset has sample sizes ranging from 1,000 to 10,000 for each patient. The proportion of the $CD3^+CD4^+CD8\beta^+$ cells in each dataset ranges from $10\%$ to $30\%$, depending on the development of the GvHD. We aim at finding the subspace that is immune to biological variation between patients but predictive of GvHD. On account of the imbalanced distribution within each domain, the performance is measured in terms of the area under ROC curve (AUC) and G-Mean \cite{kubat1997addressing}, instead of classification accuracy, which is calculated as:
\begin{equation}
G-Mean=\sqrt{\frac{TP}{TP+FN}\cdot\frac{TN}{TN+FP}}.
\end{equation} 
The G-Mean is a combinatory measurement of the accuracies of the two classes. The range is from 0 to 1, and the higher it is, the better the performance is in evaluation. We use this measurement because it is extremely sensitive towards classification errors on minority instances in the severely imbalanced dataset.

To reduce the impact of quantity imbalance among domains, we sample $n_i=1000$ cells from each patient to form the original dataset we use in the experiment, maintaining the ratio of $CD3^+CD4^+CD8\beta^+$ cells to other cells. Since there are 30,000 data points leading to a large dataset, we apply FastCOIR and FastDCM to compare with SVM with parameter $M=5$. We pick 9 patients for training and the rest 21 for test. In addition, we subsample $n_i=200$ points from 1000 for each patient for the test of smaller-scale computation with SVM, COIR, DICA, and DCM. In such a setting, we select 12 for training and 18 for test. Table \ref{gvhdtable} briefly reports the average AUC, G-Mean, and their corresponding standard deviations and computation time for large data over 20 repeats with parameters $\gamma=10$ and $\epsilon$ is between $10^{-6}$ to $10^{-5}$.

\subsection{LandMine Data}

The landmine detection problem \cite{xue2007multi} is based on airborne synthetic-aperture radar (SAR) data collected from real landmines\footnote{http://www.ee.duke.edu/~lcarin/LandmineData.zip}. In this problem, there are a total of 29 sets of data collected from different landmine fields. Each data is represented as a 9-dimensional feature vector extracted from radar images, and the class label is true mine or false mine. Since each of the 29 datasets are collected from different regions that may have different types of ground surface conditions, these datasets are considered to be dominated by different distributions, among which 9 construct the source domain and 20 for the target. 

Each landmine field has roughly 500 instances with a severely imbalanced class distribution. We uniformly downsample the instances to 200 for each field to generate a smaller set for standard algorithms, meanwhile apply fast versions on the original large set. Due to the imbalanced distribution, we also use AUC and G-Mean to measure the performance. Table \ref{landminetable} records the AUC, G-Mean running time over 10 repetitions for larger scale and 20 repetitions for the small set. Parameters are $M=2$, $\gamma=0.5$, $\epsilon=10^{-6}$ for the large-scale set and $\epsilon=10^{-3}$ for the small set.

\begin{table}[!t]
\centering
\caption{AUC and G-Mean for SVM classification on LandMine Dataset}
\vspace{0.1in}
\label{landminetable}
\begin{tabular}{|c|c|c|c|}
\hline
Methods ($n_i\approx200 $) &AUC&G-Mean&Time ($s$)\\
\hline
SVM&$0.5076\pm0.0069$ &$0.1061\pm0.0704$&\\
COIR&$0.5594\pm0.0195$ &$0.3671\pm0.0564$& 45.91\\
DICA&$0.5595\pm0.0195$ &$0.3671\pm0.0563$ & 103.08\\
DCM&$\bold{0.5640\pm0.0195}$ &$\bold{0.3853\pm0.0580}$&65.32\\
\hline
Methods ($n_i\approx500$, $M=2$)&AUC&G-Mean&Time ($s$)\\
\hline
SVM&$0.5141\pm0.0053$ &$0.1677\pm0.0297$&$$ \\
FastCOIR &$0.5450\pm0.0110$&$0.3084\pm0.0411$&$194.26$ \\
FastDCM&$\bold{0.5496\pm0.0101}$ &$\bold{0.3233\pm0.0360}$&$355.52$\\
\hline
\end{tabular}
\end{table}

\begin{table}[!t] 
\centering
\caption{AUC and G-Mean for SVM classification on Oil Field Dataset}
\vspace{0.1in}
\label{oiltable}
\begin{tabular}{|c|c|c|c|}
\hline
Methods ($n_i\approx10,000$, $M=20$)&AUC&G-Mean&Time ($s$)\\
\hline
SVM&$0.5187\pm0.000$ &$0.1963\pm 0.0000$&$$ \\
FastCOIR&$0.5783\pm0.0106$ &$0.3941\pm0.0271$&$132.14$ \\
FastDCM&$\bold{0.5907\pm0.0123}$ &$\bold{0.4248\pm0.0270}$&$216.43$\\
\hline
\end{tabular}
\end{table}

\subsection{Oil Field Data}

Oil field failure prediction \cite{liu2011semi} is a real-world problem but a hard nut to crack. The dataset we use is collected from 4 oil fields. Each field has hundreds of wells, and each well possesses almost three years of historical data. Experts want to make predictions for each oil production well given historical data, that whether it fails to run normally or not. Howbeit, various geographical environments and weather conditions induce diverse characteristics between fields. Despite different data distributions, correctly identifying failure wells is vital for timely repair, reducing oil losses, and saving human, financial, and material resources. 

Since it's also an anomaly detection problem, we still adopt AUC and G-Mean to measure the performance. Three of four fields are combined with being the source domains, and the rest is the target. We flatten the dataset by ignoring well information and assign approximately 10,000 data points to each field. To present the result more directly and persuasively, we only report the fast version performances operated on the original large dataset since sometimes distributions would vary in the sampling subspace. Due to the field-specific characteristics, we test two of the fields while the other two always remain in the source. AUC, G-Mean, and computation time are reported in Table \ref{oiltable}. Parameters are $M=20$, $\gamma=0.1$, $\epsilon=10^{-4}$.

\section{Conclusion}
In this paper, we address a nontrivial domain generalization problem by finding a central subspace in which domain-based covariance is minimized while the functional relationship is simultaneously maximally preserved. We propose a novel variance measurement for multiple domains to minimize the difference between conditional distributions across domains with solid theoretical demonstration and supports; meanwhile, the algorithm preserves the functional relationship via maximizing the variance of conditional expectations given output. Furthermore, we also provide a fast implementation that requires much less computation and smaller memory for large-scale matrix operations, suitable for not only domain generalization but also other kernel-based eigenvalue decompositions. To show the practicality of the proposed method, we compare our methods against some well-known dimension reduction and domain generalization techniques on both synthetic data and real-world applications. We show that we can achieve better quantitative results for small-scale datasets, indicating better generalization performance over unseen test datasets. The proposed fast implementation maintains the quantitative performance for large-scale problems but at a substantially lower computational cost.

\bibliographystyle{IEEEtran}
\bibliography{report}

\end{document}